\documentclass[12pt]{spieman}  % 12pt font required by SPIE;
\usepackage{amsmath,amsfonts,amssymb}
\usepackage{graphicx}
\usepackage{setspace}
\usepackage{tocloft}
\usepackage{makecell}

\usepackage[super, sort&compress]{natbib}
\setcitestyle{citesep={,}}

\usepackage{multirow}
\usepackage{hyperref}

\usepackage[utf8]{inputenc}
\usepackage{tcolorbox}
\usepackage{lipsum} % for placeholder text

\definecolor{lightbeige}{HTML}{eceae0} % Light beige color
\definecolor{darkbeige}{HTML}{b8b09a}  % Darker beige color for the frame
\usepackage{booktabs}
\usepackage{xcolor}
\usepackage{pifont}
\usepackage[figuresright]{rotating}
\newcommand{\npj}[1]{\textcolor{black}{#1}}

\title{Large Language Models for Disease Diagnosis: A Scoping Review}

\author[1, \#]{Shuang Zhou}
\author[2, \#]{Zidu Xu}
\author[3, \#]{Mian Zhang}
\author[4, \#]{Chunpu Xu}
\author[5]{Yawen Guo}
\author[6]{Zaifu Zhan}
\author[7]{Yi Fang}
\author[8]{Sirui Ding}
\author[4]{Jiashuo Wang}
\author[4]{Kaishuai Xu}
\author[9]{Liqiao Xia}
\author[1]{Jeremy Yeung}
\author[10]{Daochen Zha}
\author[11]{Dongming Cai}
\author[12]{Genevieve B. Melton}
\author[1]{Mingquan Lin}
\author[1, *]{Rui Zhang}
\affil[1]{Division of Computational Health Sciences, Department of Surgery, University of Minnesota, Minneapolis, MN, USA}
\affil[2]{School of Nursing, Columbia University, New York, New York, USA}
\affil[3]{Erik Jonsson School of Engineering and Computer Science, University of Texas at Dallas, Richardson, TX, USA}
\affil[4]{Department of Computing, The Hong Kong Polytechnic University, Hong Kong, Hong Kong SAR}
\affil[5]{Department of Informatics, University of California, Irvine, Irvine, CA, USA}
\affil[6]{Department of Electrical and Computer Engineering, University of Minnesota, Minneapolis, MN, USA}
\affil[7]{Department of Computer Science, New York University (Shanghai), Shanghai, CN}
\affil[8]{Bakar Computational Health Sciences Institute, University of California San Francisco, San Francisco, CA, USA}
\affil[9]{Department of Industrial and Systems Engineering, The Hong Kong Polytechnic University, Hong Kong, Hong Kong SAR}
\affil[10]{Department of Computer Science, Rice University, Houston, TX, USA}
\affil[11]{Department of Neurology, University of Minnesota, Minneapolis, MN, USA}
\affil[12]{Institute for Health Informatics and Division of Colon and Rectal Surgery, Department of Surgery, University of Minnesota, Minneapolis, MN, USA}

\cftpagenumbersoff{figure}
\cftpagenumbersoff{table} 
\begin{document} 
\maketitle

% Include email contact information for corresponding author
{
\noindent \footnotesize{\#}Equal contribution \\
\noindent \footnotesize\textbf{*}Correspondence: ruizhang@umn.edu
}

% {\noindent \footnotesize\textbf{*}Corresponding author: \linkable{zhan1386@umn.edu}}

\begin{spacing}{2}   % use double spacing for rest of manuscript

\begin{abstract}
\addcontentsline{toc}{section}{Abstract}
% Words Limitation: 150
Automatic disease diagnosis has become increasingly valuable in clinical practice. The advent of large language models (LLMs) has catalyzed a paradigm shift in artificial intelligence, with growing evidence supporting the efficacy of LLMs in diagnostic tasks. 
Despite the increasing attention in this field, a holistic view is still lacking. Many critical aspects remain unclear, such as the diseases and clinical data to which LLMs have been applied, the LLM techniques employed, and the evaluation methods used.
In this article, we perform a comprehensive review of LLM-based methods for disease diagnosis. 
Our review examines the existing literature across various dimensions, including disease types and associated clinical specialties, clinical data, LLM techniques, and evaluation methods.
Additionally, we offer recommendations for applying and evaluating LLMs for diagnostic tasks. 
Furthermore, we assess the limitations of current research and discuss future directions. 
To our knowledge, this is the first comprehensive review for LLM-based disease diagnosis.

\end{abstract}

\section*{Introduction}
\addcontentsline{toc}{section}{Introduction}

Automatic disease diagnosis is pivotal in clinical practice, leveraging clinical data to generate potential diagnoses with minimal human input~\cite{liu2020deepskin}. It enhances diagnostic accuracy, supports clinical decision-making, and addresses healthcare disparities by providing high-quality diagnostic services~\cite{mei2020artificial}. Additionally, it boosts efficiency, especially for clinicians managing aging populations with multiple comorbidities~\cite{li2021artificial, li2024performance, qiu2020development}. 
For example, DXplain~\cite{barnett1987dxplain} analyzes patient data to generate diagnoses with justifications. Online services also promote early diagnosis and large-scale screening for diseases like mental health disorders, raising awareness and mitigating risks~\cite{li2024performance, su2020deep, gkotsis2017characterisation, du2018extracting, caraballo2025trustworthiness}.

\npj{Advances in artificial intelligence (AI) have driven two waves of automated diagnostic systems~\cite{sajda2006machine, stafford2020systematic, kline2022multimodal, aggarwal2021diagnostic}. Early approaches utilized machine learning techniques like support vector machines and decision trees~\cite{myszczynska2020applications, fatima2017survey}. With larger datasets and computational power, deep learning (DL) models, such as convolutional, recurrent, and generative adversarial networks, became predominant~\cite{choy2023systematic, mei2020artificial, liu2020deepskin, mei2023interstitial, zhou2024open, zhou2021machine}. However, these models require extensive labeled data and are task-specific, limiting their flexibility~\cite{liu2020deepskin, hannun2019cardiologist, zhou2024open}.
The rise of generative large language models (LLMs), like GPT~\cite{NEURIPS2020_1457c0d6} and LLaMA~\cite{touvron2023llama}, pre-trained on extensive corpora, has demonstrated significant potential in various clinical applications, such as question answering~\cite{singhal2023large, yang2024cm} and information retrieval~\cite{peng2024depth, zhan2025ramie}. These models are increasingly applied to diagnostics. For example, PathChat~\cite{lu2024multimodal}, a vision-language LLM fine-tuned with comprehensive instructions, set new benchmarks in pathology. Similarly, \citet{kim2024large} reported that GPT-4 outperformed mental health professionals in diagnosing obsessive-compulsive disorder, underscoring its potential in mental health diagnostics.}

\npj{Despite growing interest, several key questions remain unresolved: Which diseases and medical data have been explored for LLM-based diagnostics (Q1)? What LLM techniques are most effective for diagnostic tasks, and how should they be selected (Q2)? What evaluation methods best assess performance of various diagnostic tasks (Q3)?
Many reviews have explored the use of LLMs in medicine~\cite{thirunavukarasu2023large, zhou2023survey, meng2024application, zhang2024data, du2024generative, wang2024survey, li2024scoping, he2023survey}, but they typically provide broad overviews of diverse clinical applications rather than focusing specifically on disease diagnosis. 
For instance, \citet{pressman2024clinical} highlighted introducing various clinical applications of LLMs, e.g., pre-consultation, treatment, and patient education.
These reviews tend to overlook the nuanced development of LLMs for diagnostic tasks and do not analyze the distinct merits and challenges in this area, revealing a critical research gap. Some reviews~\cite{omar2024utilizing, giuffre2024systematic} have focused on specific specialties—such as digestive or infectious diseases—but failed to offer a comprehensive perspective that spans multiple specialties, data types, LLM techniques, and diagnostic tasks to fully address the critical questions at hand.}

\npj{This review addresses the gap by offering a comprehensive examination of LLMs in disease diagnosis through in-depth analyses. First, we systematically investigated a wide range of disease types, corresponding clinical specialties, medical data, data modalities, LLM techniques, and evaluation methods utilized in existing diagnostic studies. Second, we critically evaluated the strengths and limitations of prevalent LLM techniques and evaluation strategies, providing recommendations for data preparation, technique selection, and evaluation approaches tailored to different contexts. Additionally, we identify the shortcomings of current studies and outline future challenges and directions. To the best of our knowledge, this is the first review dedicated exclusively to LLM-based disease diagnosis, presenting a holistic perspective and a blueprint for future research in this domain.}

\begin{figure*}
\begin{center}
\includegraphics[width = 0.99\linewidth]{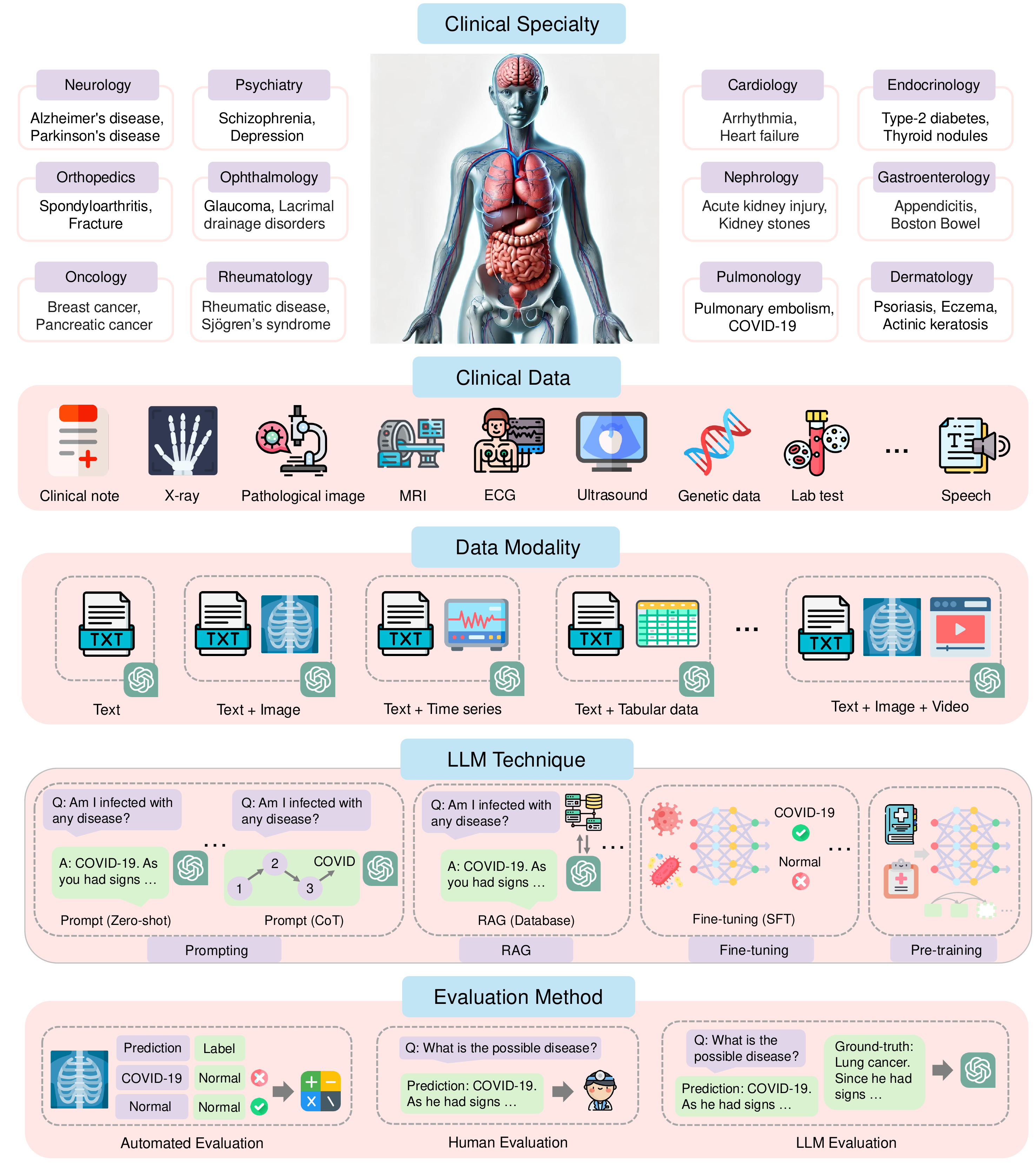}
\end{center}
\caption{Overview of the investigated scope. It illustrated disease types and the associated clinical specialties, clinical data types, modalities of the utilized data, the applied LLM techniques, and evaluation methods. \npj{We only presented part of the clinical specialties, some representative diseases, and partial LLM techniques.}}
\label{Fig1_scope_overview}
% \vspace{-0.3cm}
\end{figure*}

% Box for terms and concepts
\begin{tcolorbox}[
    colback=lightbeige,        % Light gray background
    colframe=darkbeige,         % Black border
    boxrule=0.8pt,          % Border thickness
    width=\textwidth,       % Full width of text
    title=Box 1: Terms and Concepts, % Title of the box
    fonttitle=\bfseries,    % Bold title
    coltitle=black,         % Title text color
    left=2mm,               % Padding on the left
    right=2mm,              % Padding on the right
    top=2mm,                % Padding at the top
    bottom=2mm,             % Padding at the bottom
    before skip=10pt,       % Space before the box
    after skip=10pt         % Space after the box
]

\setstretch{0.9} % Reduce line spacing for compactness
\noindent  \textbf{Disease diagnosis:} receiving clinical data, such as patient symptoms, medical history, and diagnostic tests, as input and identifying which disease explains the symptoms and signs. \\
\textbf{Diagnostic tasks:} a type of tasks that generate disease diagnoses or probability estimates for specific conditions, such as differential diagnosis and conversational diagnosis. \\
\textbf{Large language models:} a type of AI models using deep neural networks to learn the relationships between words in natural language, using large datasets of text to train. \\
\textbf{Hallucination:}  an AI-generated output that is plausible but factually incorrect or unrelated to the input, arising from limitations in training or reasoning. \\
\textbf{Prompt:} an input or instruction provided to an AI model to guide its response, often designed to elicit specific or task-relevant outputs. \\
\textbf{Chain-of-thought:} a technique enabling AI to generate multi-step reasoning by breaking down complex tasks into sequential steps for improved accuracy. \\
\textbf{Self-consistency prompt:} a method that samples diverse reasoning paths and selects the most consistent solution to enhance the reliability of outputs in reasoning tasks. \\
\textbf{Soft prompt:} a learnable embedding added to the input space of a pre-trained model to guide its behavior without modifying the model's parameters. \\
\textbf{Retrieval-augmented generation:} integrates retrieved data into LLMs, enhancing responses by leveraging external information for improved context and accuracy in content generation. \\
\textbf{Fine-tuning:} the process of adapting a pre-trained model to a specific task by training it further on a smaller, task-specific dataset. \\
\textbf{Supervised fine-tuning:} refining a pre-trained model for a task using labeled data to enhance task-specific performance. \\
\textbf{Parameter-efficient fine-tuning:} adapting pre-trained models to new tasks by updating limited parameters (e.g., adapters), reducing computational costs while preserving performance.
\textbf{Reinforcement learning from human feedback:} a method where models improve outputs by learning from human-provided feedback, aligning behavior with human goals through reinforcement learning. \\
\textbf{Pre-training:} the foundational training phase of a model on a large, general dataset to learn broad patterns, features, and representations, which can later be adapted to specific tasks through fine-tuning.
\end{tcolorbox}

\section*{Results}
\addcontentsline{toc}{section}{Results}

\subsection*{Overview of the scope}
\addcontentsline{toc}{subsection}{Overview of the scope}

\npj{This section outlines the scope of our review and key findings. Figure~\ref{Fig1_scope_overview} provides an overview of disease types, clinical specialties, data types, and modalities (Q1), and introduces the applied LLM techniques (Q2) and evaluation methods (Q3), addressing the key questions. Our analysis spans 19 clinical specialties and over 15 types of clinical data in diagnostic tasks, covering modalities such as text, image, video, audio, time series, and multimodal data. 
We categorized existing works based on LLM techniques, which fall into four categories: prompting, retrieval-augmented generation (RAG), fine-tuning, and pre-training, with the latter three further subdivided.
Table~\ref{table_LLM_techniques} summarizes the taxonomy of mainstream LLM techniques. 
Figure~\ref{Fig_flowchart} illustrates the associations between clinical specialties, modalities of utilized data, and LLM techniques in the included papers.
Additionally, Figure~\ref{Fig2_meta_analysis} presents a meta-analysis, covering publication trends, widely-used LLMs for training and inference, and statistics on data sources, evaluation methods, data privacy, and data sizes. Collectively, these analyses comprehensively depict the development of LLM-based disease diagnosis.}

\subsection*{Study characteristics}
\addcontentsline{toc}{subsection}{Study characteristics}

\begin{figure*}
\begin{center}
\includegraphics[width = 0.99\linewidth]{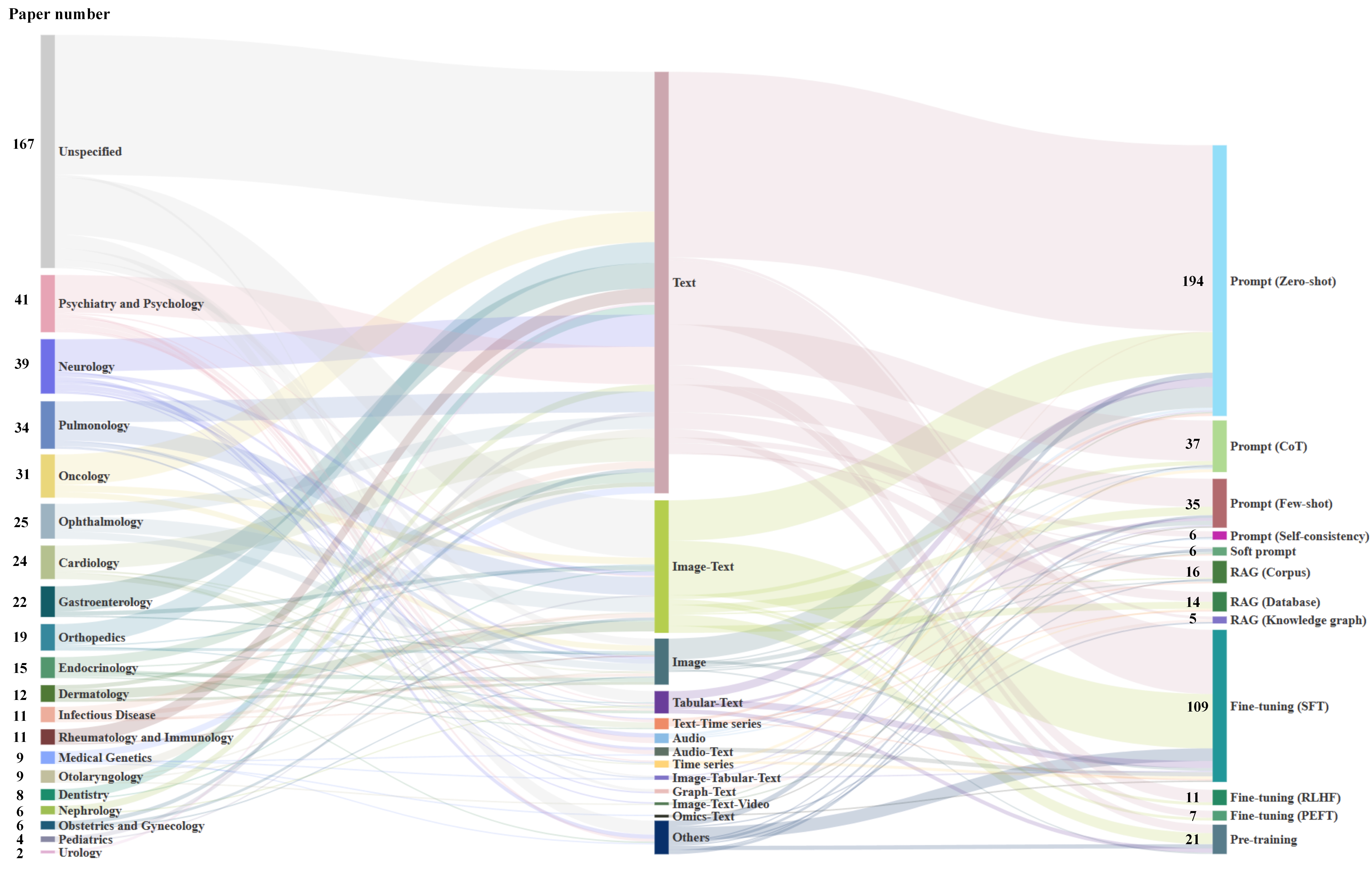}
\end{center}
\caption{Summary of the association between clinical specialties (left), data modalities (middle), and LLM techniques (right) across the included studies on disease diagnosis.}
\label{Fig_flowchart}
% \vspace{-0.3cm}
\end{figure*}

\begin{figure*}
\begin{center}
\includegraphics[width = 0.99\linewidth]{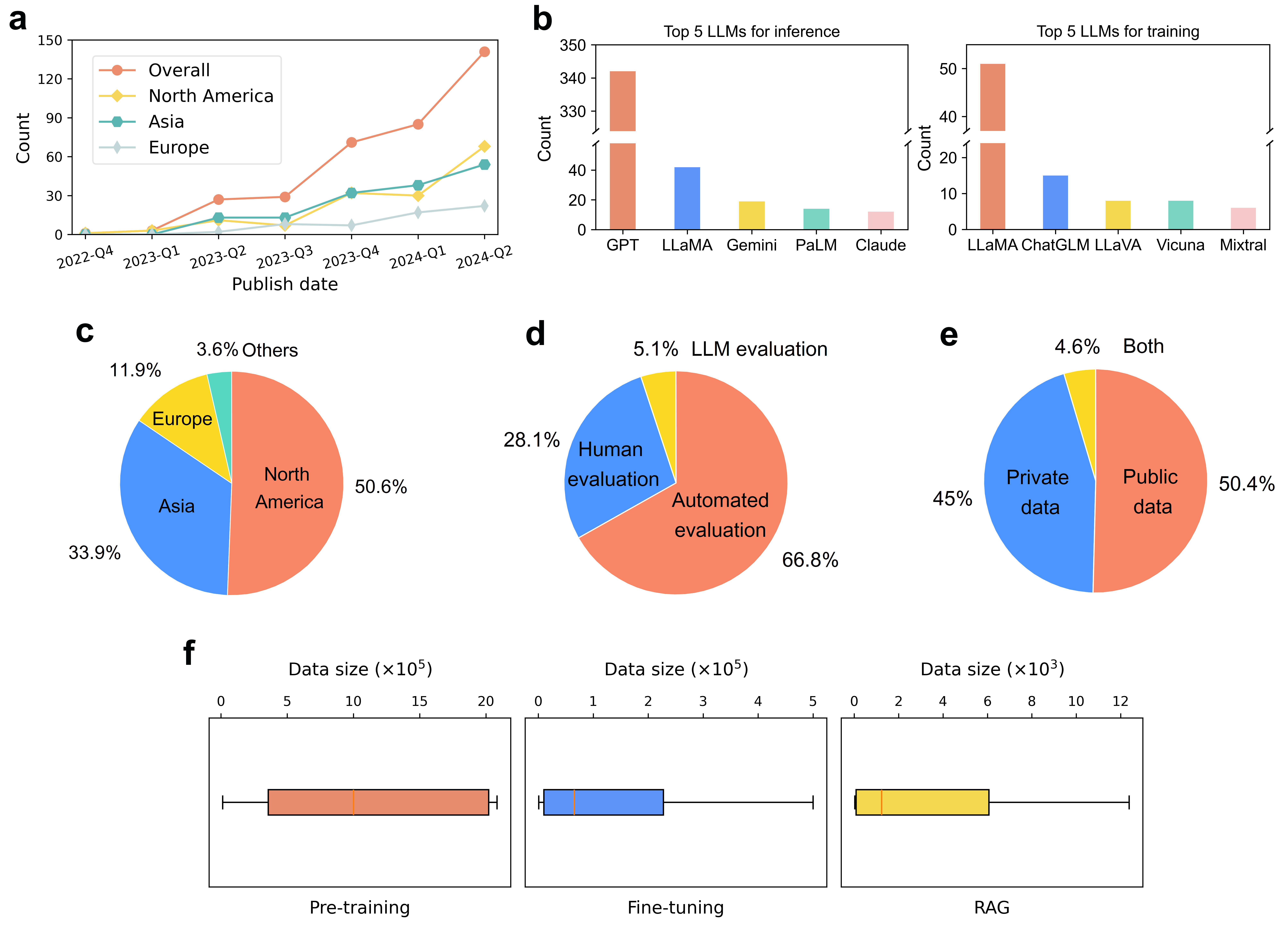}
\end{center}
\caption{Metadata of information from LLM-based diagnostic studies in the scoping review. \textbf{a} Quarterly breakdown of LLM-based diagnostic studies. Since the information for 2024-Q3 is incomplete, our statistics only cover up to 2024-Q2. \textbf{b} The top 5 widely-used LLMs for inference and training. \textbf{c} Breakdown of the data source by regions.
\textbf{d} Breakdown of evaluation methods (note that some papers utilized multiple evaluation methods). \textbf{e} Breakdown of the employed datasets by privacy status. \npj{\textbf{f} Distribution of data size used for LLM techniques. The red line indicates the median value, while the box limits represent the interquartile range (IQR) from the first to third quartiles. Notably, pre-trained diagnostic models were often followed by other LLM techniques (e.g., fine-tuning), yet this figure only includes studies that primarily used fine-tuning or RAG. Statistics for prompting methods are not included because: (\textit{i}) hard prompts generally utilize zero or very few demonstration samples, and (\textit{ii}) although soft prompts require more training data, the number of relevant studies is insufficient for meaningful distribution analysis.}}
\label{Fig2_meta_analysis}
% \vspace{-0.3cm}
\end{figure*}

\npj{As shown in Figure~\ref{Fig_flowchart}, the included studies span all 19 clinical specialties, and some specialties receive particular attention, such as pulmonology and neurology. While most studies leveraged text modality, multi-modal data, such as text-image~\cite{mai2024} and text-tabular data~\cite{kraljevic2024foresight}, are widely adopted for diagnostic tasks.
Another observation is that various LLM techniques have been applied to diagnostic tasks, and all have been used with multi-modal data (Table~\ref{table_LLM_techniques}). 
Additionally, we find an increasing number of LLM-based diagnostic studies all over the world, reflecting the field’s growing significance (Fig.~\ref{Fig2_meta_analysis}a). Among these studies, GPT~\cite{NEURIPS2020_1457c0d6} and LLaMA~\cite{touvron2023llama} families dominate inference tasks, while LLaMA and ChatGLM~\cite{glm2024chatglm} are commonly adopted for model training (Fig.~\ref{Fig2_meta_analysis}b). Fig.~\ref{Fig2_meta_analysis}c shows that most datasets originate from North America (50.6\%) and Asia (33.9\%), and 50.4\% of the studies used public datasets (Fig.~\ref{Fig2_meta_analysis}e). Evaluation methods vary: 66.8\% rely on automated evaluation, 28.1\% on human assessment, and 5.1\% on LLM-based evaluation (Fig.~\ref{Fig2_meta_analysis}d). Fig.~\ref{Fig2_meta_analysis}f reveals that the included studies employed large datasets (e.g., $5 \times 10^{5}$ samples) for pre-training diagnostic models, surpassing those primarily using fine-tuning or RAG. This phenomenon aligns with another observation that over half of pre-training models used data from multiple specialties.}

\begin{table}[]
\centering
\caption{Overview of LLM techniques for diagnostic tasks.}
\resizebox{0.99\linewidth}{!}{
\begin{tabular}{|c|l|l|}
\hline
\textbf{Techniques}           & \multicolumn{1}{c|}{\textbf{Types}} & \multicolumn{1}{c|}{\textbf{Representative studies}} \\ \hline
\multirow{5}{*}{Prompting}    & Zero-shot                          & Text~\cite{wu2024,mizuta2024}, image~\cite{noda2024,overgaard2024}, audio~\cite{hu2024exploiting, rezaii2024artificial}, text-image~\cite{benary2023}, text-time series~\cite{liu2023BioSignal, liu2023large}, text-tabular~\cite{slack2023tablet} \\ \cline{2-3} 
                              & Few-shot                           & Text~\cite{dou2023plugmed,yang2024cm}, image~\cite{siepmann2024}, text-image~\cite{mai2024,xia2024cares}, text-image-tabular~\cite{bae2024ehrxqa} \\ \cline{2-3} 
                              & CoT                                & Text~\cite{wada2024, moallem2024automated}, audio \cite{chen2023empowering}, time series~\cite{englhardt2024classification}, text-image~\cite{parth2024,busch2024} \\ \cline{2-3} 
                              & Self-consistency                   & Text~\cite{abdullahi2024learning}, audio~\cite{lim2024erd}, text-image-tabular-time series~\cite{yuhan2024} \\ \cline{2-3} 
                              & Soft prompt                        & Text~\cite{peng2024improving} , image~\cite{wenshuo2023}, tabular-time series~\cite{niu2024,belyaeva2023multimodal}, text-image-graph\cite{PENG2024103225} \\ \hline
\multirow{3}{*}{RAG}          & Knowledge graph                    & Text~\cite{wen2023mindmap}, text-time series~\cite{Zhu2024EMERGEIR} \\ \cline{2-3} 
                              & Corpus                             & Text~\cite{Ranjit2023RetrievalAC,Ge2023DevelopmentOA}, text-image~\cite{upadhyaya2024360,Xia2024RULERM}, text-time series~\cite{Yu2024ECGSI} \\ \cline{2-3} 
                              & Database                           & Text~\cite{Rau2023ACC,shi2023retrieval}, text-image~\cite{rifat2024respiratory} \\ \hline
\multirow{3}{*}{Fine-tuning}  & SFT                                & \makecell[l]{Text~\cite{Toma2023-zn,Ting2024-de,Vithanage2024-yl}, text-image~\cite{Liu2023-kf,liu2023medical,Song2024-cw}, text-video~\cite{yang2024advancing,Zhang2024-ti}, text-audio~\cite{sun2024pathasst,liu2023stone}, text-tabular~\cite{kraljevic2024foresight,slack2023tablet}} \\ \cline{2-3} 
                              & RLHF                               & Text~\cite{Dou2024-sx,wang2023clinicalgpt,Zhang2023-bv}, text-image~\cite{zhou2024large} \\ \cline{2-3} 
                              & PEFT                               & Text~\cite{Sun2024-ud,Yang2024-vr,Toma2023-zn}, text-image~\cite{Chen2024-wc} \\ \hline
Pre-training                  & -                                  & Text~\cite{Rajashekar2024-gd,Labrak2024-ib,Yang2024-vr}, text-image~\cite{liu2023medical,Chen2024-mo,Xu2024-bo}, text-tabular~\cite{Ding2024-ux,slack2023tablet}, text-video~\cite{liu2023stone}, text-omics~\cite{Xu2024-bo} \\ \hline
\end{tabular}
}
\parbox{\textwidth}{\scriptsize \textit{Note:} SFT = supervised fine-tuning, RLHF = reinforcement learning from human feedback, PEFT = parameter-efficient fine-tuning.}
\label{table_LLM_techniques}
\end{table}

\subsection*{Prompt-based disease diagnosis}
\addcontentsline{toc}{subsection}{Prompt-based disease diagnosis}

A customized prompt typically includes four components: instruction (task specification), context (scenario or domain), input data (data to process), and output indicators (desired style or role). In this review, over 60\% (N=278) of studies employed prompt-based techniques, categorized as hard prompts and soft prompts. Hard prompts are static, interpretable, and written in natural language. The most common methods included zero-shot (N=194), Chain-of-Thought (CoT) (N=37), and few-shot prompting (N=35). Among them, CoT prompting excels in thoroughly digesting input clinical cues in manageable steps to make a coherent diagnosis decision. Particularly, in differential diagnosis tasks, CoT reasoning allows the LLM to sequentially analyze medical images, radiology reports, and clinical history, generating intermediate outputs that lead to a holistic decision, with an accuracy of 64\%~\cite{busch2024}. Self-consistency prompting was used in a few studies (N=4). For instance, a study combined self-consistency with CoT prompting to improve depression prediction by synthesizing diverse data sources through multiple reasoning paths. This hybrid approach reduced the mean absolute error by nearly 50\% compared to standard CoT methods~\cite{yuhan2024}.

In contrast, soft prompts (N=6) are continuous vector embeddings trained to adapt the behavior of LLMs for specific tasks~\cite{zhangyang2023}. These prompts effectively integrate external knowledge, such as medical concept embeddings and clinical profiles, making them well-suited for complex diagnostic tasks requiring nuanced analysis. This knowledge-enhanced approach achieved F1 scores exceeding 0.94 for diagnosing common diseases like hypertension and coronary artery disease and demonstrated superiority in rare disease diagnosis~\cite{niu2024}.

Most prompt-based studies (N=221) focused on unimodal data, predominantly text (N=171). Clinical text sources like clinical notes~\cite{chung2024large}, imaging reports~\cite{delsoz2023, fink2023, moallem2024automated}, and case reports~\cite{benary2023, reese2023limitations} were commonly used. These studies often prompted LLMs with clinical notes or case reports to predict potential diagnoses~\cite{sarangi2024radiological, wang2024augmented, du2024enhancing, haider2024evaluating}. A smaller subset (N=19) applied prompt engineering to medical image data, analyzing CT scans~\cite{siepmann2024}, X-rays~\cite{PENG2024103225, xu2023elixr}, MRI scans~\cite{gertz2024potential, siepmann2024}, and pathological images~\cite{ono2024, dai2021transmed} to detect abnormalities and provide evidence for differential diagnoses~\cite{upadhyaya2024360, noda2024, ono2024, antaki2024}.

With the advancement of multimodal LLMs, 57 studies explored their application in disease diagnosis through prompt engineering. Visual-language models (VLMs) like GPT-4V, LLaVA, and Flamingo (N=37) integrated medical images (e.g., radiology scans) with textual descriptions (e.g., clinical notes)~\cite{peng2023development, suh2024comparing, pugliese2023artificial}. For example, incorporating ophthalmologist feedback and contextual details with eye movement images significantly improved GPT-4V's diagnostic accuracy for amblyopia~\cite{upadhyaya2024360}.

Beyond image-text data, more advanced multimodal LLMs (e.g., GPT-4o and Gemini-1.5 Pro) have also integrated other data types to support disease diagnosis in complex clinical scenarios. Audio and video data have been used to diagnose neurological and neurodegenerative disorders, such as autism~\cite{hu2024exploiting,deng2024hear} and dementia~\cite{rezaii2024artificial, PENG2024103225}. Time-series data, such as ECG signals and wearable sensor outputs, were used to support arrhythmia detection~\cite{liu2023BioSignal, Yu2023ZeroShotED}.  With the integration of tabular data such as user demographics~\cite{wu2023multimodal,feng2023large}, and lab test results~\cite{niu2024, ma2024clibench},
the applications have been extended to depression and anxiety screening~\cite{yuhan2024}. Omics data has been integrated to aid in identifying rare genetic disorders~\cite{liang2024genetic} and diagnose Alzheimer's disease~\cite{feng2023large}. Some studies further enhanced diagnostic capabilities by integrating medical concept graphs to provide a richer context for conditions such as neurological disorders~\cite{PENG2024103225}.

\subsection*{Retrieval-augmented LLMs for diagnosis}
\addcontentsline{toc}{subsection}{Retrieval-augmented LLMs for diagnosis}

To enhance the accuracy and credibility of the diagnosis, alleviate hallucination issues, and update LLMs' stored medical knowledge without needing re-training, recent studies~\cite{Thompson2023LargeLM,shi2023retrieval, wen2023mindmap} have incorporated external medical knowledge into diagnostic tasks. 
The external knowledge primarily comes from corpus~\cite{Thompson2023LargeLM,kresevic2024optimization,Yu2024ECGSI,upadhyaya2024360,Ghersin2023ComparativeEO,Ge2023DevelopmentOA,Xia2024RULERM,Ranjit2023RetrievalAC,zhenzhu2024gpt}, databases~\cite{shi2023retrieval,abdullahi2024learning,rifat2024respiratory,Ferber2024AutonomousAI,Yu2023ZeroShotED,Soong2023ImprovingAO,Rau2023ACC}, and knowledge graph~\cite{wen2023mindmap,Zhu2024EMERGEIR}, in the included papers.
Based on the data modality, these RAG-based studies can be roughly categorized into text-based, text-image-based, and time-series-based augmentations. 

% In text-based RAG, the majority of research~\cite{shi2023retrieval,kresevic2024optimization,Soong2023ImprovingAO,Ge2023DevelopmentOA,Ferber2024AutonomousAI,Rau2023ACC,Ghersin2023ComparativeEO} has adopted a basic retrieval strategy. In this approach, external knowledge is encoded into vector representations using sentence transformers (e.g., OpenAI's text-embedding-ada-002), which serve as retrieval sources.
% Queries were similarly encoded, allowing the system to identify and fetch the most relevant knowledge by calculating the similarity between query vectors and source vectors. This combined information was then fed into LLMs using specially designed prompts to generate diagnostic results.
In text-based RAG, most studies~\cite{shi2023retrieval,kresevic2024optimization, Soong2023ImprovingAO,Ge2023DevelopmentOA, Ferber2024AutonomousAI, Rau2023ACC, Ghersin2023ComparativeEO} utilized basic retrieval methods where external knowledge was encoded as vector representations using sentence transformers, such as OpenAI's text-embedding-ada-002. Queries were similarly encoded, and relevant knowledge was retrieved based on vector similarities. 
The retrieved data was then input into LLMs with specific prompts to produce diagnostic outcomes. In contrast, \citet{zhenzhu2024gpt} developed guideline-based GPT agents for retrieving and summarizing content related to diagnosing traumatic brain injury. They found that these guideline-based GPT-4 agents significantly outperformed the off-the-shelf GPT-4 in terms of accuracy, explainability, and empathy evaluation. Similarly, \citet{Thompson2023LargeLM} employed regular expressions to extract relevant knowledge for diagnosing pulmonary hypertension, achieving about a 20\% improvement compared to structured methods. Additionally, \citet{wen2023mindmap} integrated knowledge graph retrieval with LLMs to enable diagnostic inference by combining implicit and external knowledge, achieving an F1 score of 0.79.
% Different from previous studies where only one LLM was utilized for diagnosis, \citet{Wang2024BeyondDD} employed several LLMs, each of which was equipped with specific medical knowledge, for joint diagnosis.

% In text-image data processing, a common approach~\cite{rifat2024respiratory,upadhyaya2024360, Wu2024DiagnosisAF, Ferber2024AutonomousAI, Panagoulias2024DermacenAA} involves extracting features from input images, converting these features into textual descriptions, and subsequently applying text-based enhancement techniques. For instance, \citet{Ferber2024AutonomousAI} employed advanced models like GPT-4V to extract critical information from images to facilitate the retrieval of relevant documents in oncology diagnosis. Similarly, \citet{Ranjit2023RetrievalAC} employed multimodal models to directly compute similarities between image and text features for document retrieval. Notably, two studies fine-tuned LLMs using the retrieved documents to enhance diagnostic accuracy~\cite{Sharma2024CXRAgentVM, Xia2024RULERM}.

In text-image data processing, a common approach \cite{Ferber2024AutonomousAI, Ranjit2023RetrievalAC} involved extracting image features and text features and aligning them within a shared semantic space. For instance, \citet{Ferber2024AutonomousAI} used GPT-4V to extract crucial image data for oncology diagnostics, achieving a 94\% completeness rate and an 89.2\% helpfulness rate. Similarly, \citet{Ranjit2023RetrievalAC} utilized multimodal models to compute image-text similarities for chest X-ray analysis, leading to a 5\% absolute improvement in the BERTScore metric. 
Notably, one study fine-tuned LLMs with retrieved documents to enhance X-ray diagnostics~\cite{Xia2024RULERM}, attaining an average accuracy of 0.86 across three datasets.

For time-series RAG, most studies focused on the electrocardiogram (ECG) analysis~\cite{Yu2024ECGSI, Yu2023ZeroShotED}.  For example, \citet{Yu2024ECGSI} transformed fundamental ECG conditions into enhanced text descriptions by utilizing relevant information for ECG analysis, resulting in an average AUC of 0.96 across two arrhythmia detection datasets. 
Additionally, \citet{chen2025large} integrated retrieved disease records with ECG data to facilitate the diagnosis of hypertension and myocardial infarction.

% readmission prediction should be excluded.
% Additionally, \citet{Zhu2024EMERGEIR} employed the RAG method for predicting mortality and readmission based on multimodal EHR, attaining an AUC of 0.88 for mortality prediction and an AUC of 0.8 for readmission prediction.

\subsection*{Fine-tuning LLMs for diagnosis}
\addcontentsline{toc}{subsection}{Fine-tuning LLMs for diagnosis}
% (Primarily) What are the mainstream methods for fine-tuning-based diagnosis (i.e., the application scenarios)?
% (Briefly) What does the data used for fine-tuning LLMs look like? 
% In other words, if users plan to conduct fine-tuning, how to prepare the data?

Fine-tuning an LLM typically encompasses two pivotal stages: supervised fine-tuning (SFT) and reinforcement learning from human feedback (RLHF). 
SFT trains models on task-specific instruction-response pairs, enabling it to interpret instructions and generate outputs across diverse modalities. This phase establishes a foundational understanding, ensuring the model processes inputs effectively. RLHF further refines the model by aligning its behavior with human preferences. Using reinforcement learning, the model is optimized to produce responses that are helpful, truthful, and aligned with societal and ethical standards~\cite{Askell2021-bj}.

In medical applications, SFT enhances in-context learning, reasoning, planning, and role-playing capabilities, improving diagnostic performance. This process integrates inputs from various data modalities into the LLM’s word embedding space. For example, following the LLaVA approach~\cite{Liu2023-tn}, visual data is converted into token embeddings using an image encoder and projector, then fed into the LLM for end-to-end training.
In this review, 49 studies focused on SFT using medical texts, such as clinical notes~\cite{Toma2023-zn}, medical dialogues~\cite{wu2024medkp,He2024-hf,Xu2024-jb}, or reports~\cite{yang2024advancing, He2024-al, Chen2024-wc}. Additionally, 43 studies combined medical texts with images, including X-rays~\cite{liu2023radiology,yang2024advancing,Alkhaldi2024-hp,Lee2023-cv}, MRIs~\cite{Lee2023-cv,kwon2024large,Chen2024-wc}, or pathology images~\cite{Xu2024-bo,Zhou2023-ec,sun2024pathasst}. A few studies explored disease detection from medical videos~\cite{yang2024advancing,Zhang2024-ti}, where video frames were sampled and converted into visual token embeddings. Generally, effective SFT requires collecting high-quality, diverse responses to task-specific instructions to ensure comprehensive training.

%~\cite{liu2023medical,Li2024-vp,Zhou2024-lp,Chen2024-mo,Nisar2024-tc,Sun2024-ik,Zhao2023-zp,Blankemeier2024-ne,Xu2024-bo}
%~\cite{Toma2023-zn,Yang2023-jv,Labrak2024-ib,wang2023clinicalgpt,Zhang2023-bv,Ye2023-mg,Chen2024-tp,Guo2024-gf,yang2024mentallama}

%~\cite{Toma2023-zn,Ting2024-de,Vithanage2024-yl,Liu2024-mv,Chen2024-mf,UnknownUnknown-ya,Vithanage2024-yl,Rajaganapathy2024-gd,Yang2023-jv,Lilli2024-ju,Labrak2024-ib,Mitchell2024-ai,Xu2024-jb,Dou2024-sx,Zhang2024-ow,Yang2024-vr,Ren2024-gb,wang2023clinicalgpt,Zhang2023-bv,wu2024medkp,He2024-hf,chen2024eyegpt,Xie2024-om,Ye2023-mg,Sun2024-ud,Wang2024-gu,Liu2024-yf,Wang2024-gu,Liu2023-sa,Guo2024-gf,Basit2024-qr,Kanzawa2024-bh,Zoller2024-xc,Shmatko2024-rl,Jiang2023-oi,Nguyen-Wenker2023-zx,Chen2024-tp,yang2024mentallama,Qi2023-fl}

%~\cite{Liu2023-kf,liu2023medical,Song2024-cw,Zhang2023-zg,kang2024wolf,Li2024-vp,Gao2023-nl,Chen2024-me,Alkhaldi2024-hp,He2024-al,Zhou2024-lp,Zhou2024-oi,Ji2024-fd,Huang2024-zs,Zhang2024-fm,Chen2024-wc,Du2024-vk,Chen2024-mo,zhou2024large,Qu2023-ll,Nisar2024-tc,Kim2024-iq,Sun2024-ik,lu2024multimodal,Zhou2023-ec,Zhong2023-ht,Chen2023-wt,Lee2023-cv,Wang2024-gu,Thawkar2023-hx,Chen2024-oa,Hyland2023-cw,Zhao2023-zp,Campanella2023-cs,Yao2024-wv,Blankemeier2024-ne,Xu2024-bo}

RLHF methods are categorized as online or offline. Online RLHF, integral to ChatGPT's success~\cite{Ouyang2022-ic}, involves training a reward model on datasets of prompts and human preferences and using reinforcement learning algorithms like Proximal Policy Optimization (PPO)~\cite{Schulman2017-rh} to optimize the LLM. Studies have shown its potential in improving medical LLMs for diagnostic tasks~\cite{zhou2024large, wang2023clinicalgpt, Zhang2023-bv}. For instance, \citet{Zhang2023-bv} aligned their model with physician characteristics, achieving strong performance in disease diagnosis and etiological analysis; the diagnostic performance of their model, HuatuoGPT, surpassed GPT-3.5 in over 60\% of cases of Meddialog~\cite{zeng-etal-2020-meddialog}. However, online RLHF's effectiveness depends heavily on the reward model's quality, which may suffer from over-optimization~\cite{Gao2022-fj} and data distribution shifts~\cite{Marks2023-gv}. Additionally, reinforcement learning often faces instability and control challenges~\cite{Henderson2017-ev}.
Offline RLHF, such as Direct Preference Optimization (DPO)~\cite{Rafailov2023-jm}, frames RLHF as optimizing a classification loss, bypassing the need for a reward model. This approach is more stable and computationally efficient, proving valuable for aligning medical LLMs~\cite{Ye2023-mg, Yang2024-vr}. \citet{Yang2024-vr} reported significant performance drops on pediatric benchmarks when the offline RLHF phase was omitted. A high-quality dataset of prompts and human preferences is essential for online RLHF reward model calibration~\cite{Guo2017-rq} or the convergence of offline methods like DPO~\cite{Tajwar2024-sx}, whether sourced from experts~\cite{Ouyang2022-ic} or advanced AI models~\cite{Bai2022-rw}.

Since full training of LLMs is challenging due to high GPU demands, parameter-efficient fine-tuning (PEFT) reduces the number of tunable parameters. The most common PEFT method, Low-Rank Adaptation (LoRA)~\cite{Hu2021-zm}, introduces trainable rank decomposition matrices into each layer without altering the model architecture or adding inference latency. In this review, all PEFT-based studies (N=7) used LoRA to reduce training costs~\cite{Chen2024-wc, Yang2024-vr, Toma2023-zn}.

\subsection*{Pre-training LLMs for diagnosis}
\addcontentsline{toc}{subsection}{Pre-training LLMs for diagnosis}
% (Primarily) What are the mainstream methods for pre-training-based diagnosis (i.e., the application scenarios)?
% (Briefly) What does the data used for pre-training LLMs look like?
% In other words, if users plan to conduct pre-training, how to prepare the data?

Pre-training medical LLMs involves training on large-scale, unlabeled medical corpora to develop a comprehensive understanding of the structure, semantics, and context of medical language. Unlike fine-tuning, pre-training enables the acquisition of extensive medical knowledge, enhancing generalization to unseen cases and improving robustness across diverse diagnostic tasks.
In this review, five studies performed text-only pretraining on the LLMs from different sources~\cite{Rajashekar2024-gd, Yang2022-zj, Labrak2024-ib, Wang2024-hr}, such as clinical notes, medical QA texts, dialogues, and Wikipedia. Moreover, eight studies injected medical visual knowledge into multimodal LLMs via pretraining~\cite{liu2023medical, Xie2023-xh, Ding2024-ux, Phan2024-gz, Xu2024-bo, Chen2024-mo}. 
For instance, \citet{Chen2024-mo} employed an off-the-shelf multimodal LLM to reformat image-text pairs from PubMed into VQA data points for training their diagnostic model. 
To improve the quality of the image encoder, pretraining tasks like reconstructing images at tile-level or slide-level~\cite{Xu2024-bo}, and aligning similar images or image-text pairs~\cite{liu2023medical} are common choices.

\begin{table}[]
\centering
\caption{Overview of evaluation metrics for diagnostic tasks. Since diagnostic tasks might include explanations alongside the predicted diagnosis, existing studies also evaluated these explanatory descriptions. \npj{We categorized the metrics based on their application scenarios: G denotes that the metric requires ground-truth diagnosis for evaluation, while T indicates those applicable to textual descriptions (e.g., generated explanations). Notably, we only present a selection of representative diagnostic tasks from the included papers: disease diagnosis (DD), differential diagnosis (DDx), conversational diagnosis (CD), medical image classification (MIC), risk prediction (RP), mental health disorder detection (MHD), and diagnostic report generation (DRG).}}
\resizebox{0.99\linewidth}{!}{
\begin{tabular}{|c|c|l|c|l|}
\hline
\textbf{Type}                              & \textbf{Evaluation   metric} & \multicolumn{1}{c|}{\textbf{Purpose}}                                                         & \textbf{\npj{Scenario}} & \multicolumn{1}{c|}{\textbf{\npj{Representative task}}}       \\ \hline
\multirow{18}{*}{Automated evaluation}       & Accuracy \cite{zhang2024generalist}                    & The ratio of all   correct predictions to the total predictions                                                   & G & DD~\cite{hu2024designing},DDx~\cite{wu2023large},CD~\cite{yang2023performance},RP~\cite{Chen2024-mf},DRG~\cite{liu2023radiology},MHD~\cite{hayati2022depression}                              \\ \cline{2-5} 
                                           & Precision\cite{wang2024augmented}                    & The ratio of true   positives to the total number of positive predictions             & G & DD~\cite{wang2024augmented},CD~\cite{Liu2024-mv},MIC~\cite{busch2024},RP~\cite{Chen2024-mf},DRG~\cite{liu2023radiology}                                \\ \cline{2-5} 
                                           & Recall\cite{wang2024augmented}                       & The ratio of true   positives to the total number of actual positive cases                    & G & DD~\cite{wang2024augmented},CD~\cite{Liu2024-mv},RP~\cite{Chen2024-mf},DRG~\cite{liu2023radiology}                                 \\ \cline{2-5} 
                                           & F1~\cite{liu2023medical}                            & Calculated as the harmonic mean of precision and recall                                       & G & DD~\cite{wang2024augmented},DDx~\cite{gao2023large},CD~\cite{Liu2024-mv},MIC~\cite{sushil2024comparative},RP~\cite{Chen2024-mf},DRG~\cite{liu2023radiology}                                 \\ \cline{2-5} 
                                           & AUC\cite{zhang2023knowledge}                           & The area under the   Receiver Operating Characteristic curve                              & G & DD~\cite{PENG2024103225},CD~\cite{kotelanski2023methods},MIC~\cite{qu2023rise},RP~\cite{Chen2024-mf},DRG~\cite{liu2023radiology},MHD~\cite{bartal2024chatgpt}                              \\ \cline{2-5} 
                                           & AUPR  \cite{du2024ret}                      & The area under the   precision-recall curve                                                          & G & DD~\cite{Blankemeier2024-ne},MIC~\cite{du2024ret},RP~\cite{Acharya2024-xx},DRG~\cite{Blankemeier2024-ne}                              \\ \cline{2-5} 
                                           & Top-k accuracy\cite{tu2024towards}               & The ratio of instances with the true label in the top k predictions to total instances      & G & DD~\cite{tu2024towards},DDx~\cite{mcduff2023towards}                             \\ \cline{2-5} 
                                           & Top-k precision\cite{xu2023elixr}              & The ratio of true positives to total positive predictions within the top k predictions         & G & DD~\cite{tu2024towards},DDx~\cite{gao2023large}                              \\ \cline{2-5} 
                                           & Top-k recall \cite{chen2021automatic}                & The ratio of true   positives within the top k predictions to actual positive cases          & G & DD~\cite{tu2024towards},DDx~\cite{gao2023large}                            \\ \cline{2-5} 
                                           & Mean square error\cite{zhang2024incorporating}            & The average of the   squared differences between predicted and actual values             & G & DD~\cite{zhang2024incorporating},RP~\cite{safranek2024automated}                              \\ \cline{2-5} 
                                           & Mean absolute error\cite{safranek2024automated}          & The average of the   absolute differences between predicted and actual values             & G & DD~\cite{zhang2024incorporating},RP~\cite{safranek2024automated}                              \\ \cline{2-5} 
                                           & Cohen's $\kappa$ \cite{pedro2024exploring}                   & Measure the   agreement between predicted score and actual score                           & G & DD~\cite{pedro2024exploring}                                  \\ \cline{2-5} 
                                           & BLUE \cite{zhou2024large}                        & Calculate precision by matching n-grams between reference and generated text & T & DD~\cite{Ren2024-xo},CD~\cite{weng2023large},MIC~\cite{panagoulias2024evaluating},DRG~\cite{zhou2024large} \\ \cline{2-5} 
                                           & ROUGE \cite{dou2023plugmed}                       & Calculate F1-score   by matching n-grams between reference and generated text                & T & DD~\cite{Ren2024-xo},CD~\cite{dou2023plugmed},MIC~\cite{panagoulias2024evaluating},DRG~\cite{zhou2024large}                  \\ \cline{2-5} 
                                           & CIDEr \cite{yang2024advancing}                       & Evaluate n-gram similarity, emphasizing alignment across multiple reference texts         & T & CD~\cite{yang2024advancing},MIC~\cite{liu2024systematic},DRG~\cite{chen2024ffa}                           \\ \cline{2-5} 
                                           & BERTScore \cite{wen2023mindmap}                   & Measure similarity by comparing embeddings of reference and generated text                & T & DD~\cite{hill2023chiron},DDx~\cite{zhou2024interpretable},CD~\cite{dou2023plugmed},DRG                                \\ \cline{2-5} 
                                           & METEOR  \cite{weng2023large}                     & Evaluate text   similarity by considering precision, recall, word order, and synonym matches     & T & DDx~\cite{zhou2024interpretable},CD~\cite{weng2023large},MIC~\cite{liu2024systematic},DRG~\cite{zhou2024large}                                \\ \hline
\multirow{4}{*}{Human   evaluation}       & Necessity \cite{dou2023plugmed}                  & Whether the response   or prediction assists in advancing the diagnosis                                                                      & T & CD~\cite{dou2023plugmed}                         \\ \cline{2-5} 
                                           & Acceptance  \cite{kottlors2023feasibility}               & The degree of   acceptance of the response without any revision                & T & DD~\cite{sarangi2024radiological},CD~\cite{nair2023dera}                                \\ \cline{2-5} 
                                           & Reliability \cite{yang2024mentallama}                 & The trustworthiness   of the evidence in the response or prediction                         & T & DD~\cite{chen2024eyegpt},CD~\cite{yang2024mentallama}                                \\ \cline{2-5} 
                                           & Explainability\cite{zhenzhu2024gpt}               & Whether the response   or prediction is explainable                        & T & DDx~\cite{umerenkov2023deciphering},CD~\cite{yang2023performance}                                 \\ \hline
\multirow{9}{*}{Human  or LLM evaluation} & Correctness \cite{chen2024icga}                 & Whether the response   or prediction is medically correct                                                                                     & T & DD~\cite{Xie2023-xh},DDx~\cite{wu2023large},CD~\cite{dou2023plugmed},DRG~\cite{lyu2023translating},MHD~\cite{yang2024mentallama}                                  \\ \cline{2-5} 
                                           & Consistency \cite{wu2024medkp}                 & Whether the response or prediction is consistent with the ground-truth or input              & T & DD~\cite{kwon2024large},DDx~\cite{umerenkov2023deciphering},CD~\cite{wu2024medkp},MHD~\cite{yang2024mentallama}                                    \\ \cline{2-5} 
                                           & Clarity \cite{shi2023retrieval}                     & Whether the response   or prediction is clearly clarified                                     & T & DD~\cite{Yang2023-jv},CD~\cite{jo2024assessing}                                 \\ \cline{2-5} 
                                           & Professionality\cite{yang2024mentallama}                & The rationality of   the evidence based on domain knowledge                                   & T & CD~\cite{Yang2023-jv},MHD~\cite{yang2024mentallama}                                 \\ \cline{2-5} 
                                           & Completeness\cite{dou2023plugmed}                 & Whether the response   or prediction is sufficient and comprehensive                            & T & DDx~\cite{zhou2024interpretable},CD~\cite{yang2023performance},DRG~\cite{lyu2023translating}                                  \\ \cline{2-5} 
                                           & Satisfaction\cite{guo2024comparing}                 & Whether the response   or prediction is satisfying                                            & T & CD~\cite{nair2023dera},DRG~\cite{chen2024ffa}                                 \\ \cline{2-5}
                                           & Hallucination\cite{wu2024medkp}                & Response contains   inconsistent or unmentioned information with previous context             & T & DDx~\cite{gao2023large},CD~\cite{yang2023performance},DRG~\cite{kang2024wolf}                              \\ \cline{2-5}                                         
                                           & Relevance\cite{shi2023retrieval}                    & Whether the response   or prediction is relevant to the context                               & T & CD~\cite{shi2023retrieval},DRG~\cite{kang2024wolf}                                \\ \cline{2-5}  
                                           & Coherence \cite{he2024bp4er}                   & Assess logical   consistency with the dialogue history                                       & T & CD~\cite{He2024-hf},DRG~\cite{Zhong2023-ht}                                  \\ \hline   
\end{tabular}
}
\label{table_evaluation_metrics}
\end{table}

\subsection*{Performance evaluation}
\addcontentsline{toc}{subsection}{Performance evaluation}

\begin{figure*}
\begin{center}
\includegraphics[width = 0.98\linewidth]{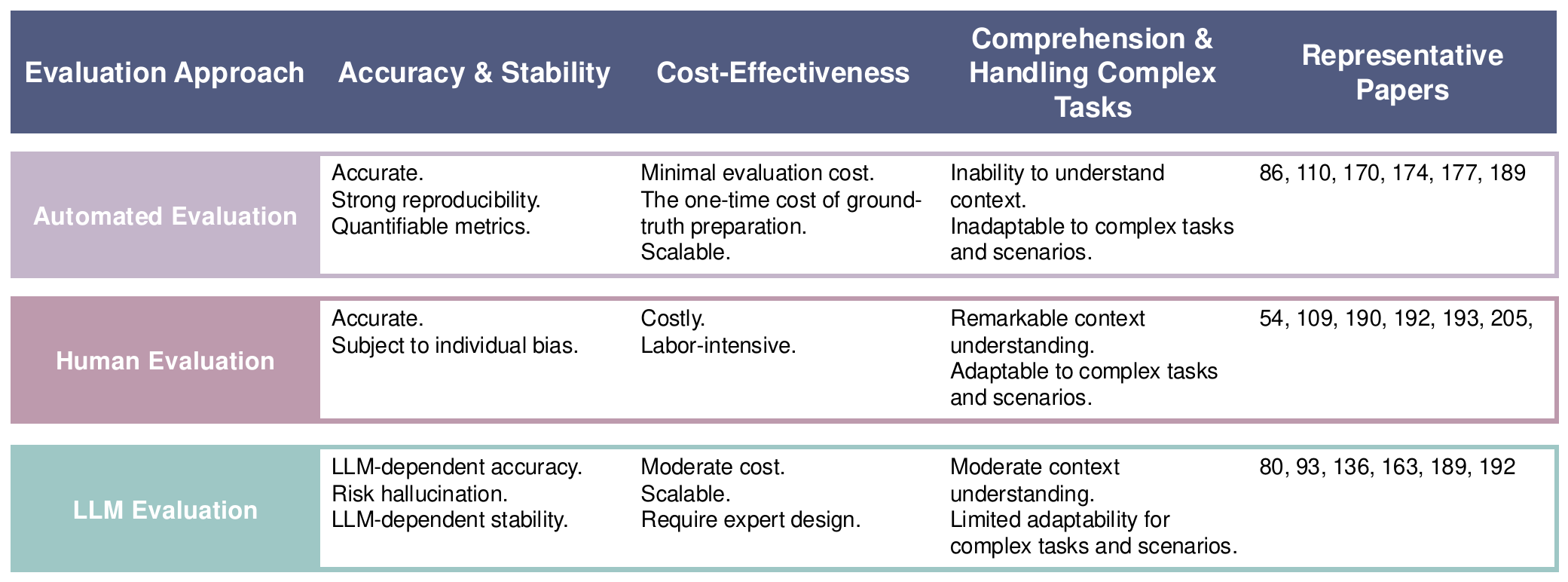}
\end{center}
\caption{\npj{Summary of the evaluation approaches for diagnostic tasks.}}
\label{Fig_evaluation}
% \vspace{-0.3cm}
\end{figure*}

\npj{Evaluation methods for diagnostic tasks generally fall into three categories (Table~\ref{table_evaluation_metrics}): automated evaluation~\cite{lu2024large}, human evaluation~\cite{lu2024large}, and LLM evaluation~\cite{li2024llms}, each with distinct advantages and limitations (Fig.~\ref{Fig_evaluation}).}

\npj{In this review, most studies (N=266) relied on automated evaluation, which is efficient, scalable, and well-suited for large datasets. These metrics can be grouped into three types. (1) Classification-based metrics, such as accuracy, precision, and recall, are commonly used for disease diagnosis. For instance, \citet{liu2023medical} evaluated COVID-19 diagnostic performance using AUC, accuracy, and F1 score.
(2) Differential diagnosis metrics, including top-k precision, assess ranked diagnosis lists. \citet{tu2024towards} employed top-k accuracy to evaluate the correctness of differential diagnosis predictions.
(3) Regression-based metrics, such as mean squared error (MSE)~\cite{safranek2024automated}, quantify deviations between predicted and actual values~\cite{zhang2024incorporating}.
Despite their efficiency, automated metrics rely on ground-truth diagnoses~\cite{zhou2024interpretable}, which may be unavailable, and cannot understand contexts, such as the readability of diagnostic explanations or their clinical utility~\cite{chen2024eyegpt}. They also struggle with complex tasks, such as evaluating the medical correctness of diagnostic reasoning~\cite{savage2024diagnostic}.}

\npj{Human evaluation (N=112), conducted by medical experts~\cite{singhal2023large, lu2024large}, does not require ground-truth labels and integrates expert judgment, making it suitable for complex, nuanced assessments. However, it is costly, time-consuming, and prone to subjectivity, limiting its feasibility for large-scale evaluation. 
Recent studies have explored using LLM evaluation (N=20), a.k.a. LLM-as-Judges~\cite{li2024llms}, to replace human experts in evaluation and combine the interpretative depth of LLM judgment with the efficiency of automated evaluation. While ground-truth accessibility is not strictly necessary~\cite{wang2023clinicalgpt, wu2024medkp}, its inclusion improves reliability~\cite{zhou2024interpretable}. Popular LLMs used for this purpose include GPT-3.5, GPT-4, and LLaMA-3. However, this approach remains constrained by LLM limitations, including susceptibility to hallucinations~\cite{wu2024medkp} and difficulties in handling complex diagnostic reasoning~\cite{li2024mediq}.}
\npj{In summary, each evaluation approach has distinct advantages and limitations, with the choice dependent on the specific requirements of the task. Figure~\ref{Fig_evaluation} guides the selection of suitable evaluation approaches for different scenarios.}

\section*{Discussion}
\addcontentsline{toc}{section}{Discussion}

\npj{This section analyzes key findings from the included studies, discusses the suitability of mainstream LLM techniques for varying resource constraints and data preparation, and outlines challenges and future research directions.}

\npj{The rapid rise of LLM-based diagnosis studies (Fig.~\ref{Fig2_meta_analysis}a) might partially be attributed to the increased availability of public datasets~\cite{fansi2022ddxplus} and advanced off-the-shelf LLMs~\cite{haider2024evaluating}. Besides, the top five LLMs used for training and inference differ significantly (Fig~\ref{Fig2_meta_analysis}b), reflecting the interplay between effectiveness and accessibility. Generally, closed-source LLMs, with their vast parameters and superior performance~\cite{zhou2024interpretable}, are favored for LLM inference, while open-source LLMs are essential for developing domain-specific models due to their adaptability~\cite{xie2024mellama}. These factors underscore the dual influence of effectiveness and accessibility on diagnostic applications.  
Additionally, the regional analysis of datasets (Fig.~\ref{Fig2_meta_analysis}c) reveals that 84.5\% of datasets originate from North America and Asia, potentially introducing racial biases in this research domain~\cite{Yang2023-jv}.}

\npj{Most studies employed prompting for disease diagnosis (Fig.~\ref{Fig_flowchart}), leveraging its advantages, such as minimal data requirements, ease of use, and low computational demands~\cite{mohammadi2024user}. Meanwhile, LLMs' extensive medical knowledge allowed them to perform competitively across diverse diagnostic tasks when effectively applied~\cite{singhal2023large, zhou2024interpretable}. For example, a study fed two data samples into GPT-4 for depression detection~\cite{tank2024depression}, and the performance significantly exceeded traditional DL-based models. 
In summary, prompting LLMs facilitates the development of effective diagnostic systems with minimal effort, contrasting with conventional DL-based approaches that require extensive supervised training on large datasets~\cite{choy2023systematic, mei2020artificial}.}

% \begin{figure*}
% % \setlength{\abovecaptionskip}{-0.2cm}
% \begin{center}
% \includegraphics[width = 0.8\linewidth]{Figures/Fig_2_technique_guideline_v2.pdf}
% \end{center}
% \caption{Summary of the advantages and limitations of the mainstream LLM techniques for diagnosis. Notably, pre-training is usually followed by fine-tuning to adapt LLMs to diagnostic tasks.}
% \label{Fig2_LLM_technique}
% % \vspace{-0.3cm}
% \end{figure*}

\npj{We then compare the advantages and limitations of mainstream LLM techniques to indicate their suitability for varying resource constraints, along with a discussion of data preparation. Generally, the choice of LLM technique for diagnostic systems depends on the quality and quantity of available data. Prompt engineering is particularly effective in few-data scenarios (e.g., zero or three cases with ground-truth diagnoses), requiring minimal setup~\cite{singhal2023large, sandmann2024systematic}.  
RAG relies on a high-quality external knowledge base, such as databases~\cite{shi2023retrieval} or corpora~\cite{kresevic2024optimization}, to retrieve accurate information during inference. Fine-tuning requires well-annotated datasets with sufficient labeled diagnostic cases~\cite{liu2023medical}. 
Pre-training, by contrast, utilizes diverse corpora, including unstructured text (e.g., clinical notes, literature) and structured data (e.g., lab results), to establish a robust knowledge foundation via unsupervised language modeling~\cite{kraljevic2024foresight, bae2024ehrxqa}. Although fine-tuning and pre-training facilitate high performance and reliability~\cite{liu2023medical}, they demand significant resources, including advanced hardware and extensive biomedical data (see Fig.~\ref{Fig2_meta_analysis}f), which are costly and often hard to obtain~\cite{singhal2023large}.}
In practice, not all diagnostic scenarios require expert-level accuracy. Applications such as large-scale screenings~\cite{hu2024designing}, mobile health risk alerts~\cite{englhardt2024classification}, or public health education~\cite{thirunavukarasu2023large} prioritize cost-effectiveness and scalability. Overall, balancing accuracy with resource constraints depends on the specific use case. 
% Figure~\ref{Fig2_LLM_technique} offers guidance for selecting the appropriate LLM technique based on available resources and diagnostic goals.

% potential findings:
% RLHF techniques are simple methods borrowed from the general domain.
% Try more advanced offline RLHF techniques.

% A smaller proportion of studies pre-trained LLMs for diagnostic tasks.
% The limited availability of large-scale, high-quality medical corpora makes pre-training LLMs specifically for diagnostic tasks challenging.
% In these models, the most frequently used medical dataset was the MIMIC-III and xxx; the most widely used pre-training technique is xxx.
% Most of the pre-trained LLMs studies demonstrated improved performance on xxx dataset. This finding aligns with previous research.
% Indeed, a primary advantage of pre-training LLMs is encoding more medical knowledge which enables them to handle rare diseases or complicated cases and expedite hallucination issues. Therefore, they can result in more accurate and reliable disease prediction, paving the way for personalized disease screening and prevention.

\begin{figure*}
\begin{center}
\includegraphics[width = 0.95\linewidth]{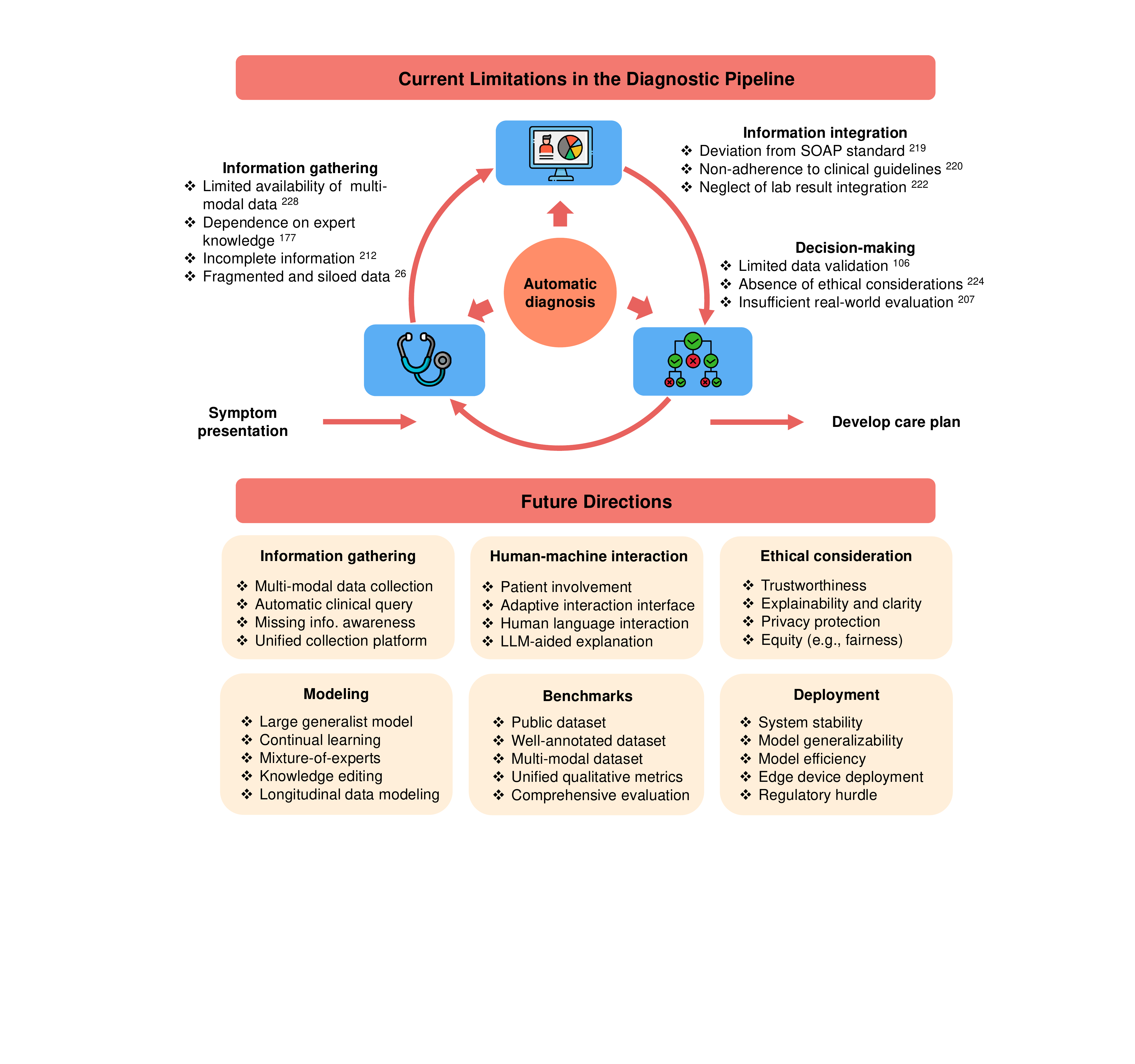}
\end{center}
\caption{Summary of the limitations and future directions for LLM-based disease diagnosis. }
\label{Fig_future_direction}
% \vspace{-0.3cm}
\end{figure*}

% One limitation lies in the limited integration of comprehensive multimodal data, such as text, images, videos, and other modalities~\cite{yuhan2024}. This contrasts with real-world diagnostic scenarios, where patient information often spans multiple modalities~\cite{saab2024capabilities}, particularly in complex, multi-organ conditions. Therefore, collecting and integrating diverse modalities is worthy of investigation.

Despite advances in LLM-based methods for disease diagnosis, this scoping review highlighted several barriers to their clinical utility (Fig.~\ref{Fig_future_direction}). 
\npj{One limitation lies in information gathering. Most studies implicitly assume that the available patient information is sufficient for diagnosis, which often fails~\cite{smith2005missing}, especially in initial consultations or with complex diseases, increasing the risk of misdiagnosis~\cite{mcinerney2024towards}. In practice, clinical information gathering is iterative, starting with initial data (e.g., subjective symptoms), refining diagnoses, and conducting further tests or screenings~\cite{adler2021next}. This process relies heavily on experienced clinicians~\cite{tu2024towards}. To reduce this dependence, recent studies have explored multi-round diagnostic dialogues to collect relevant information~\cite{shi-etal-2024-medical, sun2024conversational}. For example, AIME~\cite{tu2024towards} uses LLMs for clinical history-taking and diagnostic dialogue, while Sun et al.~\cite{sun2024conversational} utilized reinforcement learning to formulate disease screening questions. Future efforts could further embed awareness of information incompleteness into models or develop techniques for automatic diagnostic queries~\cite{zou2024aiInquiry}. Another limitation arises from the reliance on single data modalities, whereas clinicians typically synthesize information from multiple modalities for accurate diagnosis~\cite{busch2024}. Additionally, real-world health systems often operate in isolated data silos, with patient information distributed across institutions~\cite{peng2024depth}. Addressing these issues will require efforts to collect and integrate multi-modal data and establish unified health systems that facilitate seamless data sharing across institutions~\cite{zhang2025making}.
}

\npj{Barriers also exist in the information integration process. Some studies utilized clinical vignettes for diagnostic tasks without fulfilling the SOAP standard~\cite{cameron2002learning}. 
While adhering to clinical guidelines is crucial~\cite{zhang2024incorporating}, limited studies have incorporated this factor into diagnostic systems~\cite{oniani2024enhancing}. For example, \citet{kresevic2024optimization} sought to enhance clinical decision support systems by accurately explaining guidelines for chronic Hepatitis C management. 
Besides, the integration and interpretation of lab test results pose significant value in healthcare~\cite{sallam2023chatgpt}. 
For example, Bhasuran et al.~\cite{bhasuran2025preliminary} reported that incorporating lab data enhanced the diagnostic accuracy of GPT-4 by up to 30\%.
A future direction is the effective integration of lab test results into LLM-based diagnostic systems.}

\npj{Exploring clinician-patient-diagnostic system interactions offers a promising research direction~\cite{yi2024survey}. Diagnostic systems are desired to assist clinicians by providing supplementary information to improve accuracy and efficiency~\cite{siepmann2024, mcduff2023towards}, incorporating expert feedback for continuous refinement. A user-friendly interface is essential for effective human-machine interaction, enabling clinicians to input data and engage in discussions with the system. 
Human language interaction further enhances usability by allowing natural conversation with LLM-based diagnostic tools~\cite{mcduff2023towards}, reducing cognitive load. Additionally, LLM-aided explanations improve transparency by providing rationales for suggested diagnoses~\cite{savage2024diagnostic}, fostering trust, and facilitating informed decision-making among clinicians and patients.}

\npj{Most of the studies focused on diagnostic accuracy, but overlooked ethical considerations, like explainability, trustworthiness, privacy protection, and fairness~\cite{haltaufderheide2024ethics}. 
Providing diagnostic predictions alone is insufficient in clinical scenarios, as the black-box nature of LLMs often undermines trust~\cite{wu2024medkp}. Designing diagnostic models with explainability is desired~\cite{savage2024diagnostic}. For example, Dual-Inf is a prompt-based framework that offers potential diagnoses while explaining its reasoning~\cite{zhou2024interpretable}.
Besides, since LLMs suffer from hallucinations, how to enhance users' trustworthiness toward LLM-based diagnostic models is worth exploring~\cite{dou2024detection}. Potential solutions include using fact-checking tools to verify the output's factuality~\cite{tran2024leaf}.
Regarding privacy, adherence to regulations like HIPAA and GDPR, including de-identifying sensitive data, is essential~\cite{peng2024depth, Yue2020PHICONIG}. For example, SkinGPT-4, a dermatology diagnostic system, was designed for local deployment to ensure privacy protection~\cite{zhou2024pre}. Fairness is another concern, as patients should not face discrimination based on gender, age, or race~\cite{haltaufderheide2024ethics}, but research on fairness in LLM-based diagnostics remains scarce~\cite{spitale2024underneath}.}

\npj{In the context of modeling, building superior models for accurate and reliable diagnosis remains an exploration. 
While pre-training on extensive medical datasets benefits diagnostic reasoning~\cite{chen2023meditron}, many medical LLMs generally lag behind general-domain counterparts in parameter scale~\cite{yang2024mentallama, xie2024mellama}, underscoring the potential of developing large-scale generalist models for disease diagnosis.
Besides, LLMs are prone to catastrophic forgetting~\cite{peng2025continually}, where previously acquired knowledge or skills are lost when learning new information. Addressing this issue facilitates the development of generalist diagnostic models but requires incorporating robust continuous learning capabilities~\cite{yi2023towards}.
One alternative approach for accurate diagnosis involves coordinating multiple specialized models, simulating interdisciplinary clinical discussions to tackle complex cases~\cite{kim2024adaptive}. For example, Med-MoE~\cite{jiang-etal-2024-med} is a mixture-of-experts framework leveraging medical texts and images and achieved an accuracy of 91.4\% in medical image classification.
Additionally, hallucinations in LLMs undermine diagnostic reliability~\cite{dou2024detection}, necessitating solutions such as  knowledge editing~\cite{xu2024editing}, external knowledge retrieval~\cite{kresevic2024optimization}, and novel model architectures or pre-training strategies~\cite{chen2023meditron}.}
\npj{Another promising avenue is longitudinal data modeling, as clinicians routinely analyze EHRs spanning multiple years to inform decision-making~\cite{hager2024evaluation, kuratov2024babilong}. Besides, modeling temporal data helps with early diagnosis~\cite{du2024enhancing, yang2023transformehr} to improve patient outcomes. For example, early detection of lung adenocarcinoma might increase the 5-year survival rate to 52\%~\cite{huang2020machineNC}. However, challenges like irregular sampling intervals and missing data persist~\cite{cui2025timer}, necessitating advanced methodologies to effectively capture temporal dependencies~\cite{yang2024cm}.}

\npj{Another challenge in developing diagnostic models is benchmark availability~\cite{fansi2022ddxplus}. In this review, 49.6\% of the included studies relied on private datasets, which were often inaccessible due to privacy concerns~\cite{kresevic2024optimization}. Additionally, the scarcity of annotated data limits progress, as well-annotated datasets with ground-truth diagnosis enable automated evaluation, reducing reliance on human assessment~\cite{zhou2024interpretable}. Hence, constructing and releasing annotated benchmark datasets would greatly support the research community~\cite{fansi2022ddxplus}. 
Regarding performance evaluation, some studies either used small-scale data~\cite{haider2024evaluating} or unrealistic data, such as snippets from college books~\cite{savage2024diagnostic} and LLM-generated clinical notes~\cite{fansi2022ddxplus}, for disease diagnosis, while large-scale real-world data can truly validate diagnostic capabilities~\cite{hager2024evaluation}. Besides, the lack of unified qualitative metrics is another issue. For example, the evaluation of diagnostic explanation varies in different studies~\cite{dou2023plugmed, zhou2024interpretable}, including necessity~\cite{dou2023plugmed}, consistency~\cite{kwon2024large}, and compeleteness~\cite{zhou2024interpretable}. Unifying qualitative metrics foster a fair comparison. 
Additionally, many included studies failed to compare with conventional diagnostic models while recent studies reported that traditional models, e.g., Transformer~\cite{Devlin2019BERTPO}, might beat LLM-based counterparts in clinical prediction~\cite{chen2024clinicalbench}. Therefore, future studies should compare with traditional baselines for comprehensive evaluation.
}

\npj{
Regarding the deployment of diagnostic systems, several challenges warrant further investigation, including model stability, generalizability, and efficiency. Current studies have highlighted that LLMs often struggle with diagnosis stability~\cite{hager2024evaluation}, fail to generalize well across data from different institutions~\cite{zhong2023chatradio}, and encounter efficiency limitations~\cite{zhan2025epee}. For instance, even minor variations in instructions, such as from asking ``final diagnosis’’ to ``primary diagnosis’’, can drop the accuracy 10.6\% on cholecystitis diagnosis~\cite{hager2024evaluation}. Addressing these limitations will advance the reliability and applicability of diagnostic models.
Another promising avenue is deploying diagnostic algorithms on edge devices~\cite{ferrara2024large}. Such systems could enable the real-time collection of health data, such as ECG rhythms~\cite{zhou2024open}, to support continuous health monitoring~\cite{chen2025large}. However, regulatory barriers, including the stringent approval standards imposed by agencies such as the U.S. Food and Drug Administration (FDA) and the European Union’s Medical Device Regulation (MDR)~\cite{hulstaert2023gaps}, remain a significant obstacle to clinical adoption. Overcoming these challenges will be vital to ensure the safe and effective integration of LLM-based diagnostics into clinical practice.
}

In conclusion, our study provided a comprehensive review of LLM-based methods for disease diagnosis. 
Our contributions were multifaceted.
First, we summarized the disease types, the associated clinical specialties, clinical data, the employed LLM techniques, and evaluation methods within this research domain.
Second, we compared the advantages and limitations of mainstream LLM techniques and evaluation methods, offering recommendations for developing diagnostic systems based on varying user demands.
Third, we identified intriguing phenomena from the current studies and provided insights into their underlying causes.
Lastly, we analyzed the current challenges and outlined the future directions of this research field.
In summary, our review presented an in-depth analysis of LLM-based disease diagnosis, outlined its blueprint, inspired future research, and helped streamline efforts in developing diagnostic systems.

\section*{Methods}
\addcontentsline{toc}{section}{Methods}

\begin{figure*}
\begin{center}
\includegraphics[width = 0.85\linewidth]{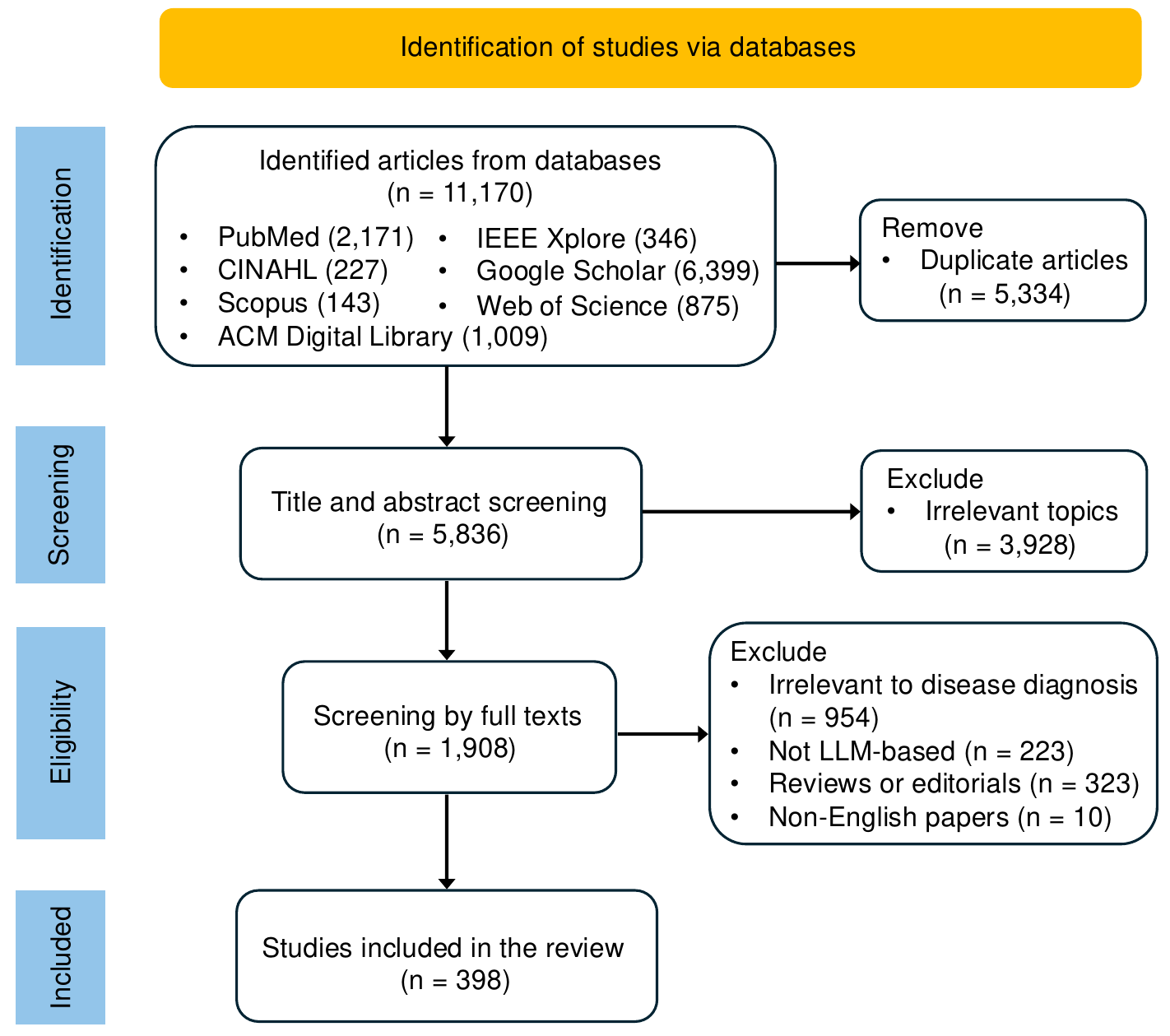}
\end{center}
\caption{PRISMA flowchart of study records. PRISMA flowchart showing the study selection process.}
\label{Fig_PRISMA}
% \vspace{-0.3cm}
\end{figure*}

\subsection*{Search strategy and selection criteria}
\addcontentsline{toc}{subsection}{Search strategy and selection criteria}
This scoping review followed the PRISMA guidelines, as shown in Figure~\ref{Fig_PRISMA}. We conducted a literature search for relevant articles published between January 1, 2019, and July 18, 2024, across seven electronic databases: PubMed, CINAHL, Scopus, Web of Science, Google Scholar, ACM Digital Library, and IEEE Xplore. Search terms were selected based on expert consensus (see Supplementary Data 1).

A two-stage screening process focused on LLMs for human disease diagnosis. 
The first stage involved title and abstract screening by two independent reviewers, excluding papers based on the following criteria: (a) articles unrelated to LLMs or foundation models, and (b) articles irrelevant to the health domain. 
The second stage was full-text screening, emphasizing language models for diagnosis-related tasks (Supplementary Data 2), excluding non-English articles, review papers, editorials, and studies not explicitly focused on disease diagnosis. 
\npj{The scope included studies that predicted probability values of diseases (e.g., the probability of depression) and the studies in which the foundation models involved text modalities (e.g., vision-language models) and utilized non-text data (e.g., medical images) as input.
Our review excluded the foundation models without text modality, such as vision foundation models, because the scope highlighted “language” models.}
\npj{Following related works~\cite{chow2024literature}, we further excluded studies purely built on non-generative language models, like BERT~\cite{Devlin2019BERTPO} and RoBERTa~\cite{Liu2019RoBERTaAR}, since the generative capability is a critical characteristic of LLMs to facilitate the development of the diagnostic system in the era of generative AI~\cite{thirunavukarasu2023large, zhou2023survey}.}
% Non-generative models or those with fewer than 400 million parameters (e.g., BERT~\cite{Devlin2019BERTPO}, RoBERTa~\cite{Liu2019RoBERTaAR}) were also omitted. 
Final eligibility was determined by at least two independent reviewers, with disagreements resolved by consensus or a third reviewer.

\subsection*{Data extraction}
\addcontentsline{toc}{subsection}{Data extraction}
Information from the articles was categorized into four groups:
(1) Basic information: title, publication venue, publication date (year and month), and region of correspondence.
(2) Data-related information: data sources (continents), dataset type, modality (e.g., text, image, video, text-image), clinical specialty, disease name, data availability (private or public), and data size.
(3) Model-related information: base LLM type, parameter size, and technique type.
(4) Evaluation: evaluation scheme (e.g., automated or human) and evaluation metrics (e.g., accuracy, precision).
See Supplementary Table 1 for the data extraction form.

\subsection*{Data synthesis}
\addcontentsline{toc}{subsection}{Data synthesis}
We synthesized insights from the data extraction to highlight key themes in LLM-based disease diagnosis. First, we presented the review scope, covering disease-associated clinical specialties, clinical data, data modalities, and LLM techniques. We also analyzed meta-information, including development trends, the most widely used LLMs, and data source distribution. Next, we summarized various LLM-based techniques and evaluation strategies, discussing their strengths and weaknesses and offering targeted recommendations. We categorized modeling approaches into four areas (prompt-based methods, RAG, fine-tuning, and pre-training), with detailed subtypes. Additionally, we examined challenges in current research and outlined potential future directions. In summary, our synthesis covered data, LLM techniques, performance evaluation, and application scenarios, in line with established reporting standards.

\section*{Data availability}
\addcontentsline{toc}{section}{Data availability}
The analyzed data are included in this article.
Aggregate data analyzed in this study will be released upon the acceptance of this paper.

%%%%% References %%%%%
\addcontentsline{toc}{section}{References}
\bibliography{reference}   
% bibliography data in reference.bib
\bibliographystyle{unsrtnat}

\section*{Acknowledgments}
\addcontentsline{toc}{section}{Acknowledgments}
This work was supported by the National Institutes of Health’s National Center for Complementary and Integrative Health under grant number R01AT009457, National Institute on Aging under grant number R01AG078154, and National Cancer Institute under grant number R01CA287413. The content is solely the responsibility of the authors and does not represent the official views of the National Institutes of Health. We also acknowledge the support from the Center for Learning Health System Sciences.

\section*{Author contributions}
\addcontentsline{toc}{section}{Author contributions}
S.Z. conceptualized the study and led the work. 
Z.Z., S.Z., J.Y., and M.Z. searched papers. 
S.Z., Z.X., M.Z., C.X., Y.G., Z.Z., S.D., J.W., K.X., Y.F., L.X., and J.Y. conducted paper screening and data extraction.
S.Z., Z.X., M.Z., and C.X. performed data synthesis and contributed to the writing. 
S.Z., Z.X., M.Z., C.X., D.Z., G.M., and R.Z. revised the manuscript.
R.Z. supervised the study.
All authors read and approved the final version.

\section*{Competing interests }
\addcontentsline{toc}{section}{Competing interests }
The authors declare no competing interests.

\end{spacing}
\end{document}